\documentclass[sigconf, authorversion]{acmart}

\AtBeginDocument{%
  }

\setcopyright{rightsretained}
\copyrightyear{2025}
\acmYear{2025}
\acmConference{SA Posters '25}{December 15-18, 2025}{Hong Kong, Hong
Kong}\acmBooktitle{SIGGRAPH Asia 2025 Posters (SA Posters '25), December 15-18, 2025} 
\acmDOI{10.1145/3757374.3771452}
\acmISBN{979-8-4007-2134-2/2025/12}

\usepackage{ccicons}
\usepackage{xspace}
\usepackage{url}

\newcommand{\commentout}[1]{}

\usepackage{color}
\usepackage{setspace}
\definecolor{Orange}{rgb}{1,0.5,0}
\definecolor{DarkGreen}{rgb}{0,0.5,0}
\definecolor{Purple}{rgb}{0.7,0,0.7}
\definecolor{Blue}{rgb}{0.2,0.2,0.8}
\definecolor{Red}{rgb}{1.0,0.0,0.0}
\definecolor{Brown}{rgb}{0.7,0.4,0.1}

\usepackage{multicol}
\usepackage{booktabs}
\usepackage{enumitem}

\usepackage{array}
\usepackage{longtable}

\begin{document}

\title{
Tactile Data Recording System for Clothing with Motion-Controlled Robotic Sliding
% Tactile Data Recording System for Quantitative Evaluation of Clothing Tactile Characteristics
}


\author{Michikuni Eguchi}
\affiliation{%
  \institution{University of Tsukuba}
  \city{Ibaraki}
  \country{Japan}
}
\affiliation{%
  \institution{Cluster Metaverse Lab}
  \city{Tokyo}
  \country{Japan}
}
\email{m.eguchi@cluster.mu}

\author{Takekazu Kitagishi}
\affiliation{%
  \institution{The University of Tokyo}
  \city{Tokyo}
  \country{Japan}
}
\affiliation{%
  \institution{ZOZO Research}
  \city{Tokyo}
  \country{Japan}
}
\email{kitagishi-takekazu588@g.ecc.u-tokyo.ac.jp}

\author{Yuichi Hiroi}
\affiliation{%
 \institution{Cluster Metaverse Lab}
 \city{Tokyo}
 \country{Japan}
}
\email{y.hiroi@cluster.mu}

\author{Takefumi Hiraki}
\affiliation{%
  \institution{University of Tsukuba}
  \city{Ibaraki}
  \country{Japan}
}
\affiliation{%
  \institution{Cluster Metaverse Lab}
  \city{Tokyo}
  \country{Japan}
}
\email{hiraki@slis.tsukuba.ac.jp}

\renewcommand{\shortauthors}{M.Eguchi et al.}

\begin{abstract}
% The feel of a garment is important in evaluating wearer comfort and is derived from the physical tactile characteristics of a finger or skin tracing the garment surface. In this study, we developed a system to record tactile data by tracing the surface of a garment with a simulated finger on the tip of a robotic arm, and recorded tactile data from 23 different garments, showing that the machine learner can classify the garment type with a certain accuracy. This result will contribute to the fundamental technology for objectively quantifying the subjective feel of clothes based on their physical characteristics.
% The tactile sensation of clothing is important for the wearer's comfort. We evaluate clothing by sliding our fingers across it, integrating the sense of movement with the tactile stimulation transmitted to the skin to judge the cloth's nuanced texture. However, existing tactile clothing databases do not include movement-related data, such as sliding speed and direction. In this study, we developed a tactile data recording system to create a multimodal haptic database of clothing with varied sliding velocities and directions. To evaluate the recorded tactile data, we conducted a clothing identification task with machine learning using data from 23 clothing items. The results showed that our recorded tactile data has diverse features that current machine learning models can easily distinguish. These results objectively quantify the subjective feel of clothing based on its physical characteristics.
The tactile sensation of clothing is critical to wearer comfort. To reveal physical properties that make clothing comfortable, systematic collection of tactile data during sliding motion is required. Recent studies demonstrate that tactile data obtained through stroking with systematically varied speeds and directions encode material properties more accurately.
However, prior methods are optimized for small samples and inadequately address non-destructive recording of intact garments, limiting their scalability to large clothing databases where non-destructive measurement is essential. We propose a robotic arm-based system for collecting tactile data from clothing. The system performs stroking measurements with a simulated fingertip while precisely controlling speeds and directions without cutting clothing, thus enabling the creation of motion-labeled, multimodal tactile databases.
Machine learning evaluation showed that including motion-related parameters (velocity and direction) improved identification accuracy for audio and/or acceleration, demonstrating the efficacy of motion-related labels for characterizing clothing tactile sensation.
% This work provides a foundation for objective quantification of clothing tactile sensations and contributes to quality investigation, classification, and tactile reproduction applications.

\end{abstract}


\begin{teaserfigure}
  \centering
  \includegraphics[width=0.96\textwidth]{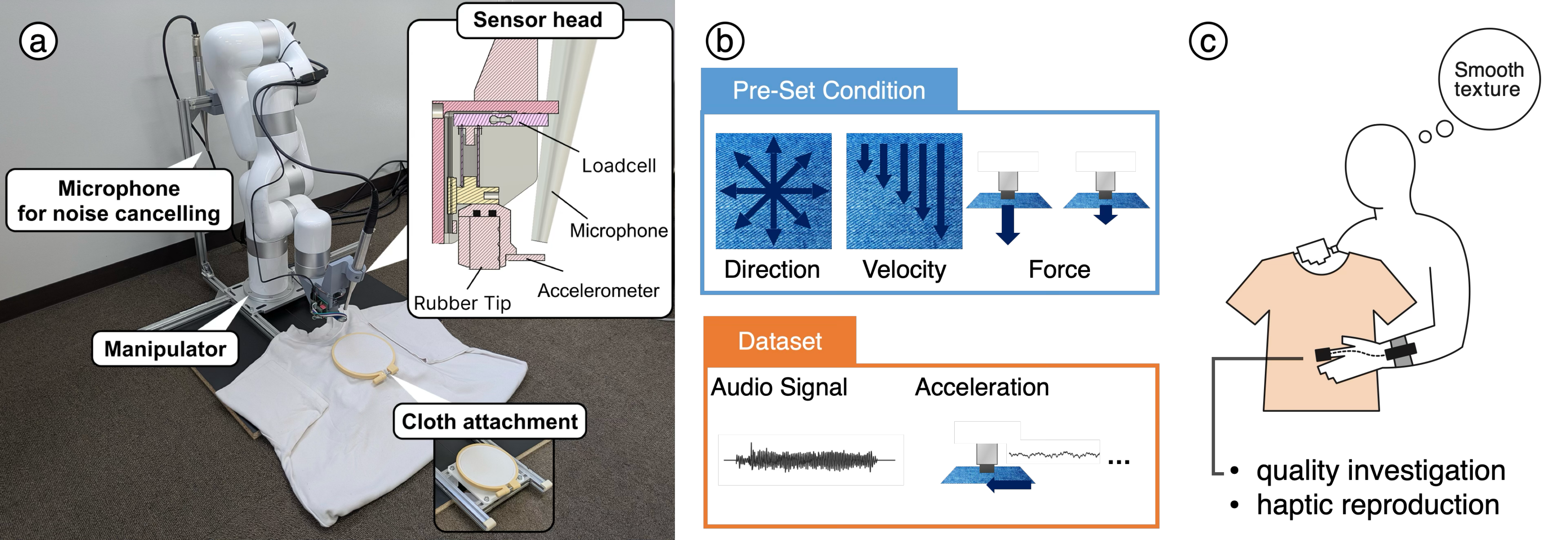}
  \caption{
    Overview of this research: (a) This system uses a robotic arm with a sensor head that has a simulated finger. The robotic arm slides the sensor head across fabric surfaces with controlled movements. The sensor head integrates two omnidirectional microphones, one triaxial accelerometer, and one load cell.
    (b) When the user places clothing on the attachment, the system automatically records multimodal tactile data under preset conditions for force applied to the garment, sliding speed, and sliding direction (8 directions, 5 speeds, and 2 kinds of force). The recorded data consists of frictional audio signals, surface images, and acceleration. 
    (c) This system can contribute to investigating clothing quality and reproducing tactile sensations of clothing.
  }
  \Description{
    The teaser figure presents a diagram of a system designed for haptic feedback and material sensing. The main components are labeled in the diagram. A "Sensor head" is depicted with several integrated sensors, including a "Loadcell" , an "Accelerometer" , and a "Microphone". The microphone is also noted as being "for noise cancelling". A "Rubber Tip"  is attached to the sensor head. The "Sensor head" is part of a "Manipulator".
    The diagram also shows the outputs from these sensors, which include "Audio Signal" and "Acceleration". The system uses a "Dataset" and has a "Pre-Set Condition" , such as "Smooth texture". The system provides "Haptic feedback based on force, velocity, direction". The outputs from the system are "Force" , "Velocity" , and "Direction". The diagram also mentions a "Cloth attachment".
  }
  \label{fig:teaser}
\end{teaserfigure}


\maketitle

\section{Introduction}
The tactile sensation of clothing is critical to wearer comfort. Consumers judge fabric texture by running their fingers over surfaces, integrating cutaneous stimulation and movement sensation~\cite{lederman2009haptic}.
From this perspective, systematic collection of tactile data during sliding motion promises to reveal physical properties that make clothing comfortable. However, existing clothing tactile databases lack systematic recording of motion variables such as sliding speed and direction, with data captured through uncontrolled finger-mounted sensors~\cite{HU20061081}. This limits precise analysis of tactility conditioned on motion factors: position, normal force, and movement direction.
Recent studies demonstrate that tactile data obtained through stroking with systematically varied speeds and directions encode material properties more accurately and improve classification performance~\cite{eguchi2025cluster}. However, these methods measure cut fabric swatches instead of intact garments. This limits their ability to be used with large clothing databases, where non-destructive measurement is essential.

We propose a robotic arm-based system for collecting tactile data from clothing (Fig.~\ref{fig:teaser}). The system performs stroking measurements with a pseudo-fingertip while precisely controlling motion without cutting clothing, enabling creation of motion-labeled, multimodal tactile databases. This provides a foundation for objective quantification of clothing tactile sensations and contributes to quality investigation, classification, and tactile reproduction applications~\cite{kitagishi2023telextiles}.

\section{Data Evaluation}
\begin{figure}[t]
    \centering
    \includegraphics[width=0.8\linewidth]{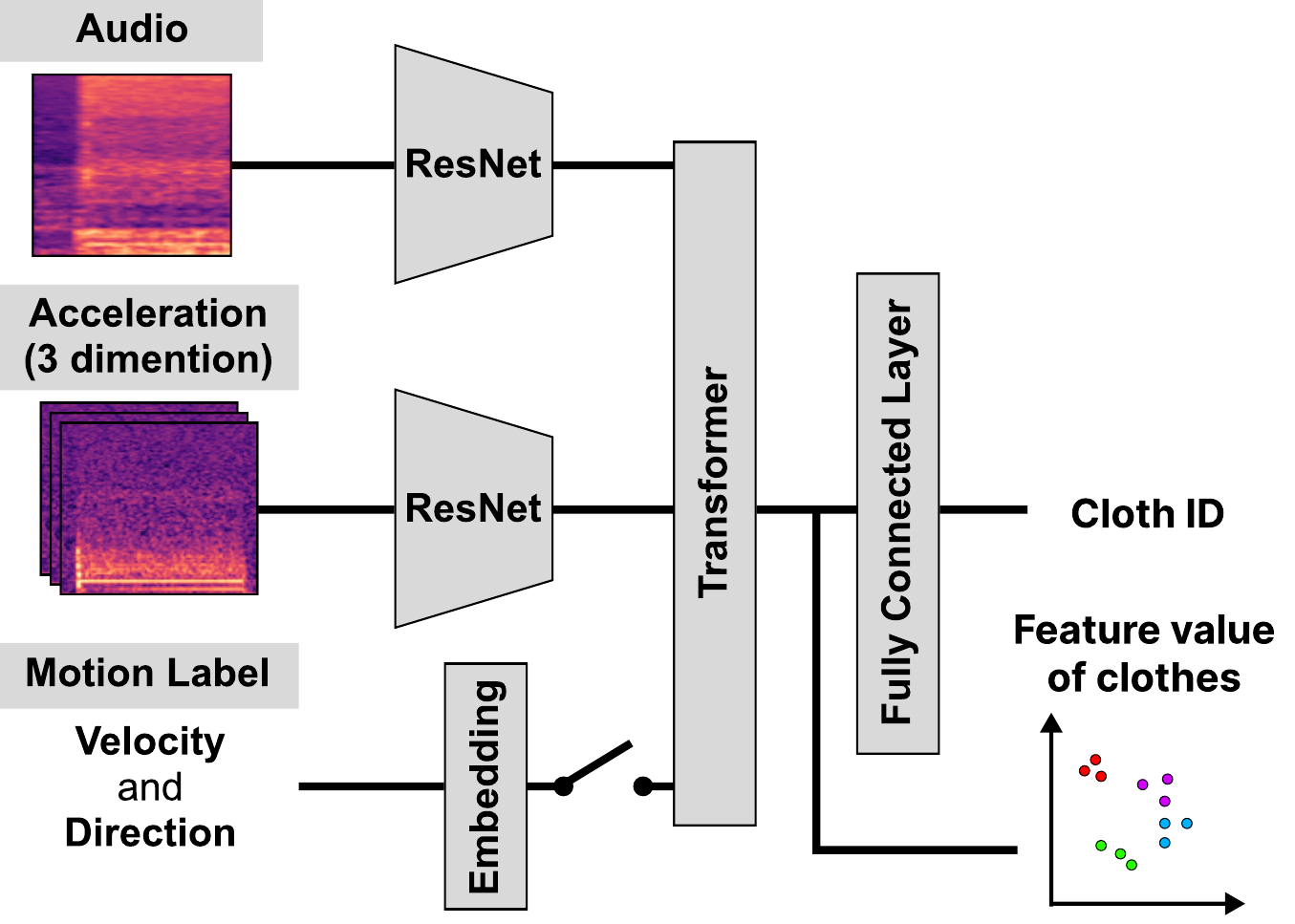}
    \vspace{-2mm}
    \caption{
        Machine learning architecture. We extracted features from audio and/or acceleration signals. We also investigated integrating motion-related parameter.
    }
    \label{fig:classifier}
\end{figure}

To assess the sufficiency of our recorded tactile data for clothing discrimination, we conducted a clothing identification task using machine learning. We utilized data from 23 distinct clothing items, varying in material (such as cotton, polyester, and wool), weave, and thickness. We then evaluated the accuracy of clothing label identification with the machine learning architecture, as shown in Fig.~\ref{fig:classifier}\commentout{, and visualized the tactile features using the highest-performing classifier (Fig.~\ref{fig:feature_value_vis})}.

\begin{table}[t]
    \caption{Clothing identification accuracy (\%). }
    \label{tab:classification_result}
    \vspace{-2mm}
    \centering
    \scalebox{1.0}{
    \begin{tabular}{c|c c c}
        \textbf{w/o Motion} & \textbf{Audio} & \textbf{Acceleration} & \textbf{Audio\&Acceleration} \\
         \hline
            with Motion & ~93.75 & ~44.57 & ~93.48 \\
            no Motion & ~87.09 & ~43.34 & ~92.80 
        \end{tabular}
        }
\end{table}

% \begin{figure}[t]
%     \centering
%     \includegraphics[width=0.8\linewidth]{supplements/cloth_texture_feature_value.pdf}
%     \caption{List of recorded garments and visualization of garment features after dimensionality reduction.
%      We selected 23 types of garments with differing materials, surface structures, thicknesses, and softness.
%     }
%     \label{fig:feature_value_vis}
% \end{figure}

Table~\ref{tab:classification_result} shows the results. It shows that including motion-related parameters, specifically velocity and direction, improved the identification accuracy for audio and/or acceleration, demonstrating the efficacy of motion-related labels on the tactile sensation of clothing. This resembles human tactile perception, where accurate perception of motion increases the accuracy of texture perception. In terms of modality comparison, the accuracy was high with audio, or with audio and acceleration, while accuracy was low with acceleration alone. Since the model can classify the tactile data of 23 different types of clothing with high accuracy, it indicates that our recorded tactile data contained diverse features that are readily distinguishable by CNN. Specifically, audio contains a significant amount of information related to clothing tactile sensation. Regarding the integration of acceleration with other modalities, when motion labels are absent, adding acceleration data significantly improved accuracy. This suggests that acceleration data plays a complementary role in providing motion information. However, when accurate motion labels already exist, the acceleration data may become redundant and could potentially act as noise.
% Table~\ref{tab:classification_result} shows the results. It shows that including motion-related parameters, specifically velocity and direction, improves the identification accuracy for audio and/or acceleration, which demonstrates the efficacy of motion-related labels on the tactile sensation of clothing. This matches the human tactile perception that, in judging the tactile sensation of clothing, humans integrate not only tactile stimulation but also motion information. A comparison of the modalities shows that accuracy is high with audio and audio\&acceleration, while it is low with acceleration alone. This indicates that our recorded tactile data contains diverse features readily distinguishable by the CNN, as it was able to classify 23 types of clothing with high accuracy. In particular, audio contains a significant amount of information related to clothing tactile sensation. Multimodal integration with acceleration proves beneficial; for the "no motion" case, adding acceleration data to the audio significantly improved accuracy, suggesting that acceleration plays a complementary role. However, when motion labels are already present, the acceleration data may become redundant and potentially act as noise, indicating that the motion parameters themselves provide a key component of the information captured by acceleration.

\section{Discussion}
This study identifies several areas for improvement. First, there is a lack of human sensory information. Future work should incorporate human validation to correlate objective tactile data with subjective ratings of clothing comfort and feel. This will help bridge the gap between physical properties and qualitative human experience. Second, the dataset should include a greater variety of clothing, such as materials with very fine or complex textures, as well as extremely soft or deformable materials. This will enhance the generalizability of our findings. Third, the variation of tactile data needs to be enriched. While our simulated finger is made of urethane rubber and can capture roughness, future data acquisition efforts should aim to capture the tactile properties of clothing, such as stiffness, thickness, and warmth.

\begin{acks}
This work was supported by JST Moonshot Research \& Development Program Grant Number JPMJMS2013 and JSPS KAKENHI Grant Number JP25K00145 and JP25H00722.

\end{acks}

\bibliographystyle{ACM-Reference-Format}
\bibliography{bib-base}

\appendix

\end{document}